\def\paperID{13350}
\def\confName{CVPR}
\def\confYear{2024}

\newif\ifreview 
\newif\ifarxiv \newcommand{\arxiv}{\arxivtrue}
\newif\ifcamera 
\newif\ifrebuttal

\arxiv

\pdfoutput=1
\documentclass[10pt,twocolumn,letterpaper]{article}
\usepackage{booktabs}
\usepackage{adjustbox}
\usepackage{enumitem}
\usepackage{tabularx}
\usepackage{amssymb}
\usepackage{graphicx} 
\usepackage{bbding}
\usepackage{amsmath}
\usepackage{titlesec}

\usepackage[T1]{fontenc}
\usepackage[utf8]{inputenc}
\usepackage{authblk}
\usepackage{titlesec}

\ifreview \usepackage[review]{cvpr} \fi
\ifarxiv \usepackage[pagenumbers]{cvpr} \fi
\ifrebuttal \usepackage[rebuttal]{cvpr} \fi
\ifcamera \usepackage{cvpr} \fi

\usepackage{graphicx}	
\usepackage{amsmath}	
\usepackage{amssymb}	
\usepackage{booktabs}
\usepackage{times}
\usepackage{microtype}
\usepackage{epsfig}
\usepackage[table,xcdraw,dvipsnames]{xcolor}
\usepackage{caption}
\usepackage{float}
\usepackage{placeins}
\usepackage{color, colortbl}
\usepackage{stfloats}
\usepackage{enumitem}
\usepackage{tabularx}
\usepackage{xstring}
\usepackage{multirow}
\usepackage{xspace}
\usepackage{url}
\usepackage{subcaption}
\usepackage{xcolor}
\usepackage[hang,flushmargin]{footmisc}

\ifcamera \usepackage[accsupp]{axessibility} \fi

% ...

\ifarxiv  \fi

\newcommand{\R}[1]{{%
    \textbf{%
        \ifstrequal{#1}{1}{\textcolor{red}{R#1}}{%
        \ifstrequal{#1}{2}{\textcolor{blue}{R#1}}{%
        \ifstrequal{#1}{3}{\textcolor{magenta}{R#1}}{%
        \ifstrequal{#1}{4}{\textcolor{teal}{R#1}}{%
                           \textcolor{cyan}{R#1}%
        }}}}%
    }%
}}

\usepackage{xr-hyper}

\makeatletter
\newcommand*{\addFileDependency}[1]{
  \typeout{(#1)}
  \@addtofilelist{#1}
  \IfFileExists{#1}{}{\typeout{No file #1.}}
}

\makeatother

\definecolor{cvprblue}{rgb}{0.21,0.49,0.74}
\usepackage[pagebackref,breaklinks,colorlinks,citecolor=cvprblue]{hyperref}
\usepackage[capitalize]{cleveref}
\crefname{section}{Sec.}{Secs.}
\crefname{table}{Table}{Tables}
\crefname{figure}{Fig.}{Figs.}

\begin{document}
	\title{You Only Need Two Detectors to Achieve Multi-Modal 3D Multi-Object Tracking}
	
	\author[1,2]{Xiyang Wang}
	\author[,1]{Chunyun Fu\thanks{Corresponding author}}
	\author[1]{Jiawei He}
	\author[1]{Mingguang Huang}
	\author[1]{Ting Meng}
	\author[2]{Siyu Zhang}
	\author[2]{Hangning Zhou}
	\author[2]{Ziyao Xu}
	\author[2]{Chi Zhang}
	\affil[1]{Chongqing University \qquad $^{2}$Mach-drive}

	\maketitle
	\begin{abstract}
 In the classical tracking-by-detection (TBD) paradigm, detection and tracking are separately and sequentially conducted, and data association must be properly performed to achieve satisfactory tracking performance. In this paper, a new end-to-end multi-object tracking framework is proposed, which integrates object detection and multi-object tracking into a single model. The proposed tracking framework eliminates the complex data association process in the classical TBD paradigm, and requires no additional training. Secondly, the regression confidence of historical trajectories is investigated, and the possible states of a trajectory (weak object or strong object) in the current frame are predicted. Then, a confidence fusion module is designed to guide non-maximum suppression for trajectories and detections to achieve ordered and robust tracking. Thirdly, by integrating historical trajectory features, the regression performance of the detector is enhanced, which better reflects the occlusion and disappearance patterns of objects in real world. Lastly, extensive experiments are conducted on the commonly used KITTI and Waymo datasets. The results show that the proposed framework can achieve robust tracking by using only a 2D detector and a 3D detector, and it is proven more accurate than many of the state-of-the-art TBD-based multi-modal tracking methods. The source codes of the proposed method are available at \url{https://github.com/wangxiyang2022/YONTD-MOT}.
  
  keywords: 3D MOT, Camera and LiDAR fusion, object detection and tracking.

\end{abstract}
	\section{Introduction}
\label{sec:intro}
\quad Multi-object tracking (MOT) is intended for determining the position and identity (i.e., trajectory ID) of objects in each frame of image or point cloud data. Depending on the sensor information involved, MOT can be divided into two major types: 2D tracking and 3D tracking. The former is commonly applied in fields such as camera-based security surveillance while the latter is suitable for applications such as autonomous driving and robot navigation using LiDAR sensors. Currently, common MOT paradigms include tracking-by-detection (TBD)\cite{03,05,10,14,32,37,42}, joint detection and embedding (JDE) \cite{35,43}, tracking-by-attention (TBA) \cite{07,22,30}, and joint detection and tracking (JDT) \cite{01,40,44}.

\begin{figure}[tp]
	\centering
	\includegraphics[width=\linewidth]{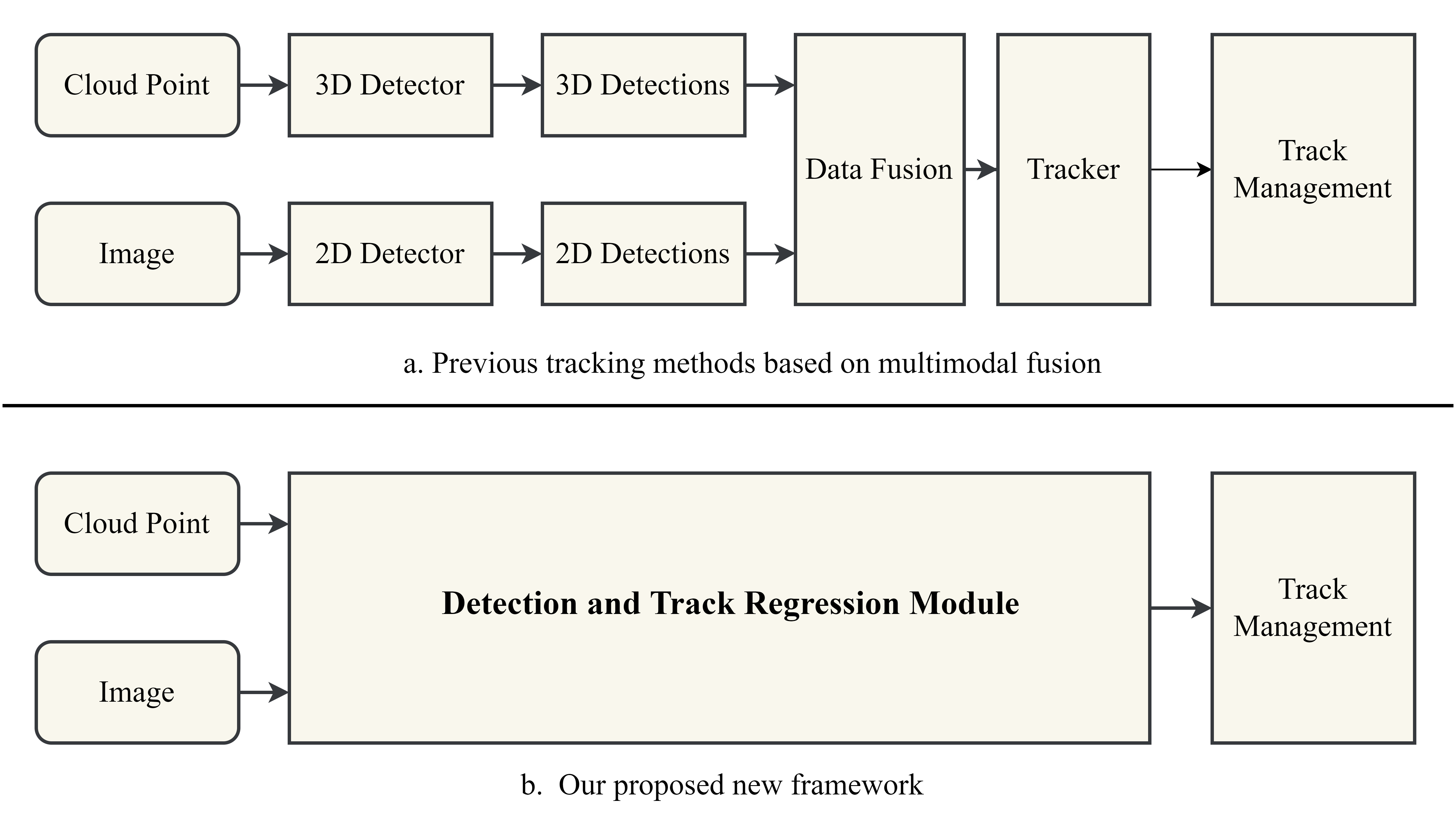}
	\caption{(a) In prior works, 2D detection and 3D detection are fused before tracking. (b) In the present work, an end-to-end MOT framework is proposed based on multi-modal fusion.}
	\label{fig:fig1}
\end{figure}

MOT methods based on a single modality often underperform those based on multi-modal fusion, due to lack of sufficient information about objects. Existing MOT methods based on multi-modal fusion commonly follow the conventional TBD paradigm \cite{19, 33, 34, 36} . Using this paradigm, 3D detectors \cite{09, 27, 28} and 2D detectors \cite{12, 24, 25} are first used to detect objects in point clouds and images respectively, and then feature fusion or post-fusion is performed to realize fusion of 3D and 2D detections. Upon completing the detection stage, a data association strategy is executed to match detections with tracks. The TBD paradigm features two distinct modules – detection and data association, and it usually involves a complex data association strategy to achieve correct matching between detections and trajectories. For this reason, it cannot be used to achieve end-to-end multi-object tracking, as demonstrated in Figure \ref{fig:fig1}.

Apart from the TBD paradigm, the JDE paradigm has also been employed in some multi-modal MOT methods, such as \cite{16,41}. Compared to the TBD paradigm, the feature extraction process in the JDE paradigm is incorporated into detectors, however a complex data association stage, similar to the TBD paradigm, is still inevitable. 

In addition to the TBD and JDE paradigms, regression-based tracking frameworks (i.e., the JDT paradigm) such as \cite{01,40} have also been proposed in the literature. These methods use two-stage 2D detectors for object detection, employ historical trajectories as proposals for regression, and then perform track-detection association in the current frame through non-maximum suppression (NMS). It must be noted that the existing regression-based methods are all 2D tracking solutions and cannot achieve 3D tracking for autonomous driving applications. Additionally, the robustness of these methods is not satisfactory because only single-modal data is used.

In this study, a novel end-to-end 3D MOT method named YONTD-MOT is proposed based on multi-modal fusion, which effectively incorporates detection and data association into one single model in the context of multi-modal sensory data. Unlike previous multi-modal fusion-based MOT methods, the proposed method incorporates the data association process into the post-processing stage of the original 3D two-stage detector. By this means, the need for a complex data association module is completely eliminated. Furthermore, regression confidence of historical trajectory is investigated to analyze the status of a trajectory, based on which a module incorporating trajectory regression confidence is proposed and used to guide the NMS process, so as to achieve ordered and robust tracking.

\indent The main contributions of this work include:  
\vspace{-0.11cm}   
\begin{itemize}
	\item	A novel end-to-end multi-modal fusion-based MOT method is proposed, which eliminates the complex data association process in the classic TBD paradigm, and allows use of a single model to perform both detection and tracking, without the need for additional detector training.
	
	\item	Regression confidence of historical trajectory is introduced to investigate the status of a trajectory, and classification is made accordingly. A confidence fusion module is then proposed to guide the NMS process, which leads to ordered and robust tracking.
	
	\item 	In the trajectory regression module, a historical trajectory feature fusion mechanism is proposed to further enhance the regression capability of the detector.
	
	\item 	The proposed MOT method achieves outstanding tracking performance on two commonly used datasets (i.e. KITTI and Waymo), in comparison with state-of-the-art (SOTA) MOT methods in the current literature. Our code is made publicly available for the benefit of the community.  
\end{itemize}

	\section{Related Works}
\label{sec:related}
\subsection{TBD-based MOT methods}

\quad The TBD paradigm was first proposed by Bewley et al. \cite{03}, and has now become a predominant MOT framework. It usually consists of two sequential steps: object detection and data association. The first step is to obtain the position information of objects in the current frame, while the second step is to match historical trajectories with current detections. Wojke et al.\cite{37} proposed an MOT solution named DeepSORT, based on the classical approach – SORT \cite{03}. In the architecture of DeepSORT, a pedestrian re-identification (re-ID) network was added to the original framework of SORT, and with this modification, appearance of objects is taken into consideration and overall tracking performance is greatly enhanced. Bochinski et al. \cite{04} proposed an MOT method suitable for high-frame-rate scenarios, which relies on two assumptions -- sufficiently high frame rate and sufficiently high detection accuracy (only one detection is produced for each object to be tracked). With these two assumptions satisfied, the commonly used ‘intersection over union (IoU)’ is employed as the similarity metric for data association. Although this method features fast tracking, its tracking robustness is limited. Zhang et al. \cite{42} pointed out that low-confidence detections should not be filtered out during tracking, and they proved that retaining low-confidence detections could produce more true positives (TPs) than false positives (FPs). In general, the TBD-based methods have dealt with detection and data association in two independent modules, and ignored correlation between them.

\titlespacing{\subsection}{0pt}{\parskip}{-\parskip}
\subsection {JDE-based MOT methods}

\quad Although the popular TBD framework generally achieves good tracking performance, it requires an additional feature extraction network to obtain appearance information of objects, which increases computational complexity of the tracker. Wang et al. \cite{35} proposed a framework named  joint detection and embedding (JDE) which integrates an object detection module and an appearance feature extraction module into a single network, having greatly improved the processing speed of MOT algorithms. However, it has been pointed out in \cite{43} that the anchor-based detector used in \cite{35} presents several limitations. On the one hand, an object may be simultaneously estimated by multiple adjacent anchors, on the other hand, it is also possible that different objects are estimated by the same anchor. In addition, down-sampling feature maps, which is commonly performed in object detection applications, can lead to difficulties for learning re-ID features. Considering the above limitations, Zhang et al. \cite{43} proposed an MOT method named FairMOT, based on the anchor-free detection method CenterNet \cite{11}. In FairMOT, two homogenous branches are employed to perform object detection task and re-ID feature extraction task respectively, having further improved the tracking performance of the JDE paradigm. It should be noted that though incorporation of some steps in data association into a detection module has been attempted in the JDE paradigm, a complex data association strategy is still required in subsequent steps.

\subsection {JDT-based MOT methods}

\quad As mentioned previously, in the TBD paradigm, detection and tracking are completed by two separate modules, which makes it impossible to achieve joint optimization of the detection and tracking modules. Bergmann et al. \cite{01} proposed a two-stage detector named Tracktor, in which bounding boxes tracked in the previous frame serve as proposals for the current frame. Regression is performed to derive positions of objects in the current frame and then non-maximum suppression (NMS) is used to realize tracking. By this means, detection and tracking are integrated into a single model. Based on Tracktor, Zhang et al. \cite{40} proposed to add an optical flow prediction branch to further improve the tracking performance. Two deep neural network modules are employed in this approach, one for learning target-wise motions using optical flows, and the other for refining and fusing targets. Nevertheless, these methods still cannot deliver highly robust tracking performance as only single-modal data is utilized, though detection and tracking have been integrated into a single module. Moreover, these methods only perform 2D tracking based on images, and are incapable of achieving 3D tracking in the world coordinate system.

In this paper, a 3D MOT framework is proposed based on multi-modal data fusion for joint detection and tracking. This framework can effectively fuse data in different modality and realize robust 3D tracking simply relying on the detection network. Besides, a historical trajectory regression confidence-guided NMS strategy for detections and trajectories is also proposed for the first time to improve the robustness of 3D MOT. The details of the proposed MOT framework are explained in the following section.

	\section{Propoesd Method}
\label{sec:method}

\begin{figure*}[t]
	\centering
	\includegraphics[width=\linewidth]{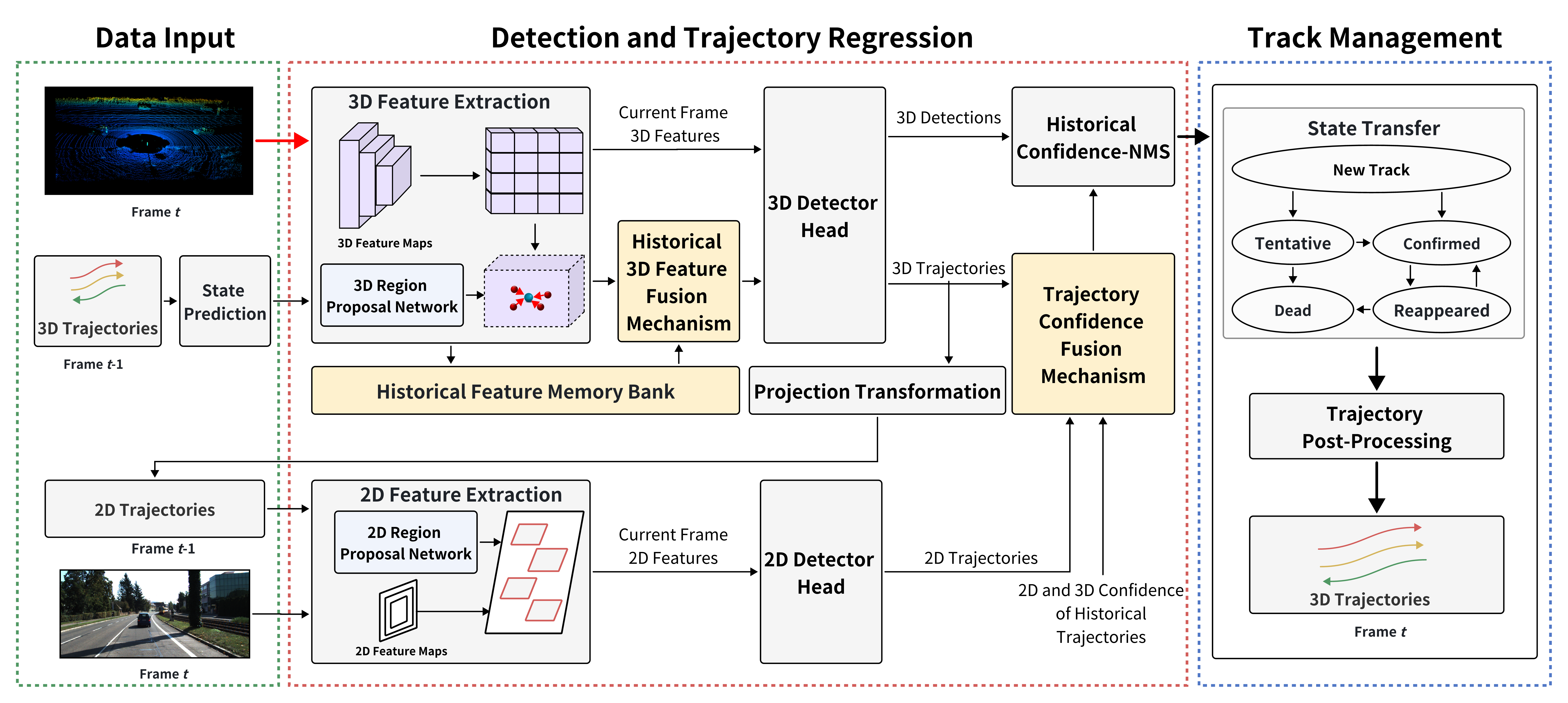}
	\caption{An overview of the proposed YONTD-MOT framework. }
	\label{fig:fig2}
\end{figure*}

\quad The multi-modal fusion-based MOT method proposed in this paper is structured as shown in Figure \ref{fig:fig2}. This structure consists of three main modules: data input module, detection and trajectory regression module, and track management module. 

The detection and trajectory regression module is further composed of several major components, including the two-stage 2D and 3D detectors, the historical feature fusion mechanism, the trajectory confidence fusion mechanism, and the NMS component based on trajectory regression confidence. First, 3D bounding boxes in the current frame are provided by a two-stage 3D detector based on the point cloud data in the current frame, and the 3D trajectory states in the previous frame are predicted and propagated to the current frame by means of a Kalman filter. Vehicle motion transformation from the previous frame to the current frame is conducted based on the GPS/IMU data, and this motion transformation is then used to compensate for the pose change of the predicted trajectories resulting from vehicle motion. Next, the predicted trajectories after motion compensation are used as proposals, which, together with the 3D features in the current and historical frames, are fed to the two-stage 3D detector for regression, so as to obtain the positions and confidences of previous trajectories in the current frame. Meanwhile, the regressed 3D trajectories are projected onto a 2D image, and the projected 2D trajectories and the obtained image in the current frame are fed to the image-based two-stage detector to obtain trajectory confidences. Then, the confidences of regressed trajectories obtained from the 3D and 2D detectors are fused with the historical confidences of trajectories. Lastly, the fused trajectory confidences are used to guide the NMS process of trajectories and detections; namely, NMS is performed for trajectories and detections in a descending order of the fused trajectory confidences, thereby achieving effective data association.

The trajectory management module is used to manage the states (e.g. dead, tentative, and confirmed) ,  lifecycle of a trajectory and trajectory post-process . In this study, we adopted the same trajectory management strategy as in DeepFusionMOT  \cite{34}, and interested readers are referred to this work for details.

\titlespacing{\subsection}{0pt}{\parskip}{-\parskip}
\subsection {Two-Stage Detectors}

\quad The proposed method requires use of two-stage detectors to acquire 2D and 3D object detections. For 2D detections, the commonly used Faster R-CNN \cite{25} was used and for 3D detections, the recently published CasA \cite{23} was employed. As a matter of fact, our proposed method does not require any specific types of detectors, i.e., any other two-stage detectors can be used. The 2D and 3D detectors were trained on the KITTI and Waymo datasets.

The employed two-stage detectors first extract candidate regions from objects through the region proposal networks, and then classify and locate the extracted regions through the regression networks. In the tracking process, the 3D detector is first used to acquire 3D detections in the current frame $t$, and we denote these detections by  $D _ { t } ^ { 3\textnormal{D} } = \{ D _ { t } ^ { 1 } , D _ { t } ^ { 2 } , \cdots , D _ { t } ^ { n } \}$. Then the 3D trajectories in the previous frame ${t - 1}$, denoted by $T _ { t - 1 } ^ { 3\textnormal{D} } = \{ T _ { t - 1 } ^ { 1 } , T _ { t - 1 } ^ { 2 } , \cdots , T _ { t - 1 } ^ { m } \}$ , are taken as proposals, which are classified and located by the 3D detector to obtain the regressed 3D trajectories in the current frame, $T _ { t } ^ { 3\textnormal{D} } = \{ T _ { t } ^ { 1 } , T _ { t } ^ { 2 } , \cdots , T _ { t } ^ { m } \}$ , and their corresponding 3D trajectory regression confidences, $S _ { t } ^ { 3\textnormal{D} } = \{  S _ { t } ^ { 1 } ,  S _ { t } ^ { 2 } , \cdots ,  S _ { t } ^ { m } \}$ . In the meantime, the regressed 3D trajectories $T _ { t } ^ { 3\textnormal{D} }$ are projected onto a 2D image to form the projected 2D trajectories $T _ { t - 1 } ^ { 2\textnormal{D} }$ , which are taken as proposals and fed to the 2D detector for classification and regression to obtain the 2D trajectory regression confidences $S _ { t } ^ { 2\textnormal{D} } = \{ S_ { t } ^ { 1 } , S _ { t } ^ { 2 } , \cdots , S _ { t } ^ { m } \}$.

\subsection {Historical Feature Fusion Mechanism}

\setlength{\parskip}{0pt}

\quad So far, almost all existing detection methods have been focused on improving the detection performance of a single shot, without taking into consideration of temporal information. Specifically, the current detection methods have overlooked the fact that in the real world objects indeed disappear gradually from the sensor field of view (FOV), and sudden disappearance of targets rarely happens. On the other hand, in the existing literature, it is a common sense that the improvement of detection performance can undoubtedly aid tracking, however, whether enhancing tracking performance can also facilitate detection has not received enough attention so far.

\begin{figure}[h]
	\centering
	\resizebox{0.45\textwidth}{!}{\includegraphics[width=\linewidth]{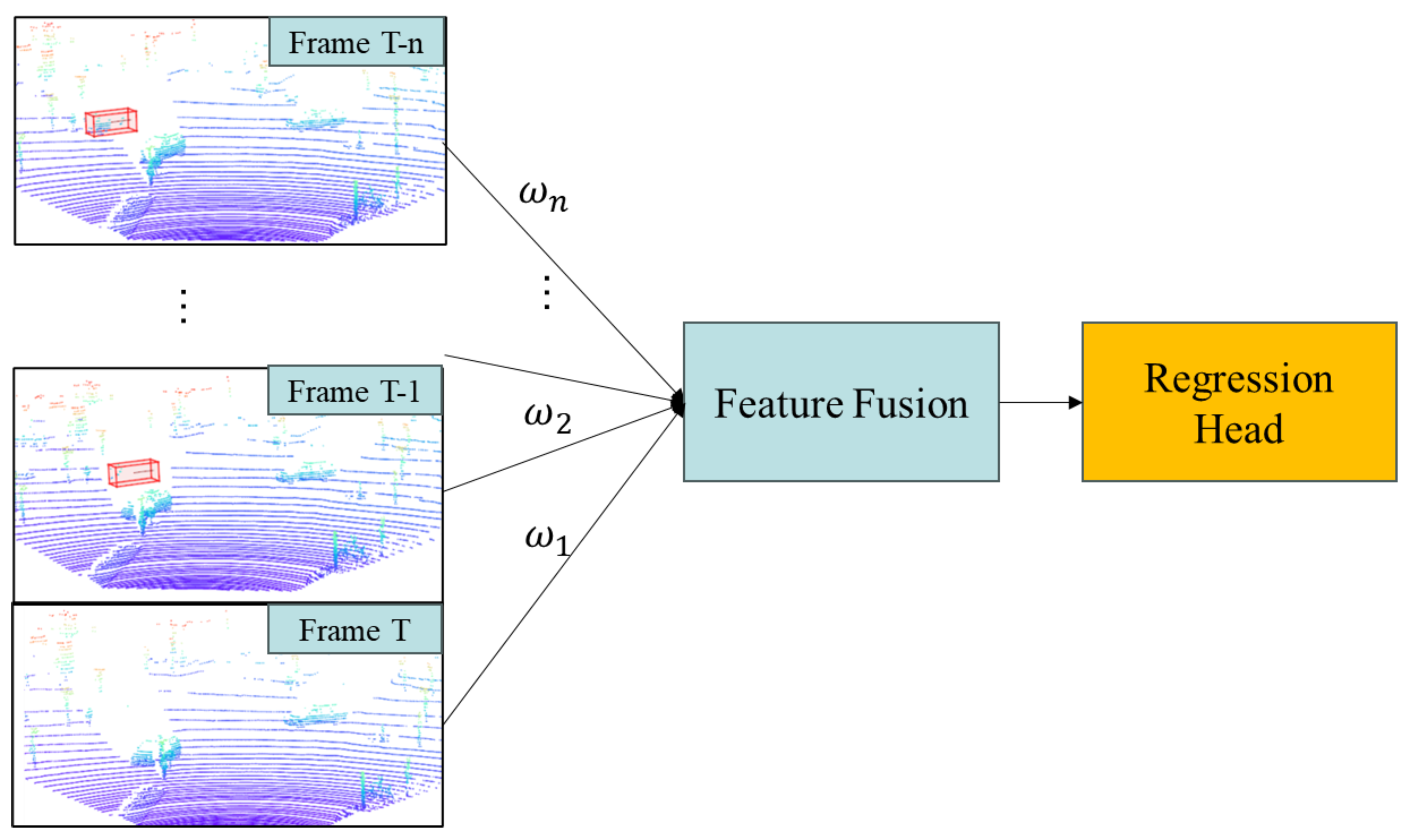}}
	\caption{Schematic of the proposed historical feature fusion mechanism. In frame t, since the red bounding box is occluded, the detector misses the detection. By fusing the historical features of this trajectory, the missing features in the current frame can be recovered, thereby effectively enhancing the detector’s performance.}
	\label{fig:fig3}
\end{figure}

As mentioned above, theoretically, an excellent detector should be able to perceive a target’s gradually leaving the sensor FOV, namely, the confidence of this gradually leaving target should also decrease in a gradual pattern. However, the widely used detection confidence in the existing literature cannot faithfully reflect the object’s motion pattern over time. In other words, when an object is gradually being occluded or leaving the sensor FOV, its detection confidence usually undergoes drastic and sudden changes, which contradicts the object’s motion pattern (i.e. gradual disappearance). To tackle this shortcoming, in this study we propose to incorporate historical trajectory information into the regression process, as shown in Figure \ref{fig:fig3}. Specifically, we cache and fuse historical features through the Historical Feature Memory Bank module based on the following formula:
\begin{equation}
	\label{eq:eq1}
	f _ { h i s } ^ { 3 d } = \frac { 1 } { n } \sum _ { k = t - n - 1 } ^ { t - 1 } ( \tilde { s } _ { k } ^ { j , 3 d } \times f _ { k } ^ { j , 3 d } ).
\end{equation}


Furthermore, the historical features are fused with the current features and then fed into a two-stage 3D detector for regression, obtaining the following corresponding regression confidence $ \tilde { s } _ { t } ^ { j , 3 d }$  in frame $t$:
\begin{equation}
	\label{eq:eq2}
	\tilde { s } _ { t } ^ { j , 3 d } = d e t e c t o r \left\{ ( 1 - s _ { t } ^ { j , 3 d } ) \times f _ { h i s } ^ { 3 d } + s _ { t } ^ { j , 3 d } \times f _ { t } ^ { j , 3 d } \right\}.
\end{equation}

\noindent where $ s _ { t } ^ { j , 3 d } $ denotes the regression confidence obtained from the feature of the $j$-th trajectory, transformed from frame $t$-1 to frame $t$. This regression confidence $ s _ { t } ^ { j , 3 d } $ characterizes the probability of the $j$-th trajectory being occluded in the current frame $t$, thereby adjusting the fusion weight with historical trajectory features. $ f _ { t } ^ { j , 3 d } $ indicates the feature of the $j$-th trajectory in frame $t$. $ \tilde { s } _ { t } ^ { j , 3 d } $ represents the regression confidence in the current frame, obtained by introducing historical features from previous $n$ frames. Fusing the historical trajectory features into the current frame can enhance the detection probability of object in the current frame. 

It’s worth mentioning that we can only fuse 3D features over time, and the same procedure cannot be applied to 2D image features. 2D object detection is performed on an image domain and no object depth information is involved, which commonly causes difficulties when dealing with occlusions. Specifically, in case of occlusion, the features extracted from the 2D proposals are likely from the occluding object, rather than the occluded object. In this case, fusing these extracted 2D features can lead to ambiguity, and in turn, performance deterioration or even failure of the detector. In comparison, the 3D detector provides object detections with depth information, and in case of occlusion, the occluding and the occluded objects can be differentiated by means of their depth difference. In other words, we are able to utilize historical 3D trajectories to infer the trajectory information in the current frame. We have found that by introducing a historical feature fusion module, tracking performance can be significantly improved, particularly effective in reducing false negatives (FN) and false positives (FP). Note that the reduction in FN can lead to improvement of recall for detection tasks. However, as the focus of this article is specifically on target tracking, we did not conduct experiments for detection performance, which will be investigated in our next step of research.

\vspace{-5pt}
\subsection {Trajectory Confidence Fusion Mechanism}

\quad In the tracking process, gradual decrease in the regression confidence of an object indicates that the object is gradually disappearing or being occluded. In this study, an object with low regression confidence is considered a “weak object”. On the contrary, an object with high regression confidence is most likely to be present and thus considered a “strong object”. For these two types of objects, different strategies should be used. Typically, “strong objects” should be matched with higher priority and “weak objects” should be matched with lower priority. 

To handle “strong objects” and “weak objects”, a regression confidence fusion module is introduced in this paper. First, 3D trajectories are input into the 3D detector for regression to obtain corresponding regression confidences. Then, for the $j$-th trajectory, its obtained regression confidence in frame $t$ is fused with the historical regression confidences in the previous $n$ frames, according to the following equation:
\begin{equation}
	\label{eq:eq3}
	\tilde { s } _ { 3 d } = \frac { 1 } { n + 1 } ( \sum _ { k = t - n - 1 } ^ { t - 1 } \tilde { s } _ { 3 d } ^ { k } + \tilde { s } _ { 3 d } ^ { t } ) .
\end{equation}

\indent Furthermore, the 2D trajectory (generated by projecting the 3D trajectory onto the 2D image) is input into the 2D detector to derive the 2D regression confidence. This confidence is then fused with previous 2D confidences of the $j$-th trajectory in the previous $n$ frames, as follows:
\begin{equation}
	\label{eq:eq4}
	s _ { 2 d } = \frac { 1 } { n + 1 } ( \sum _ { k = t - n - 1 } ^ { t - 1 } s _ { 2 d } ^ { k } + s _ { 2 d } ^ { t } ) .
\end{equation}

\noindent where $s _ { 2 d } ^ { t }$ represents the 2D regression confidence of the $j$-th trajectory in the current frame $t$, and $s _ { 2 d }  ^ { k }$ denotes the previous 2D regression confidence of the $j$-th trajectory in frame $k$.

Lastly, the above-obtained 2D confidence ($s _ { 2 d }$) and 3D confidence ($\tilde { s } _ { 3 d }$), which have taken into account historical feature information, are further fused according to the following equation to obtain the final regression confidence of the $j$-th trajectory in frame $t$:
\begin{equation}
	\label{eq:eq5}
	s _ { f } = \frac { 1 } { 2 } ( \tilde { s } _ { 3 d } + s _ { 2 d } ) .
\end{equation}

\begin{figure}[h]
	\centering
	\includegraphics[width=\linewidth]{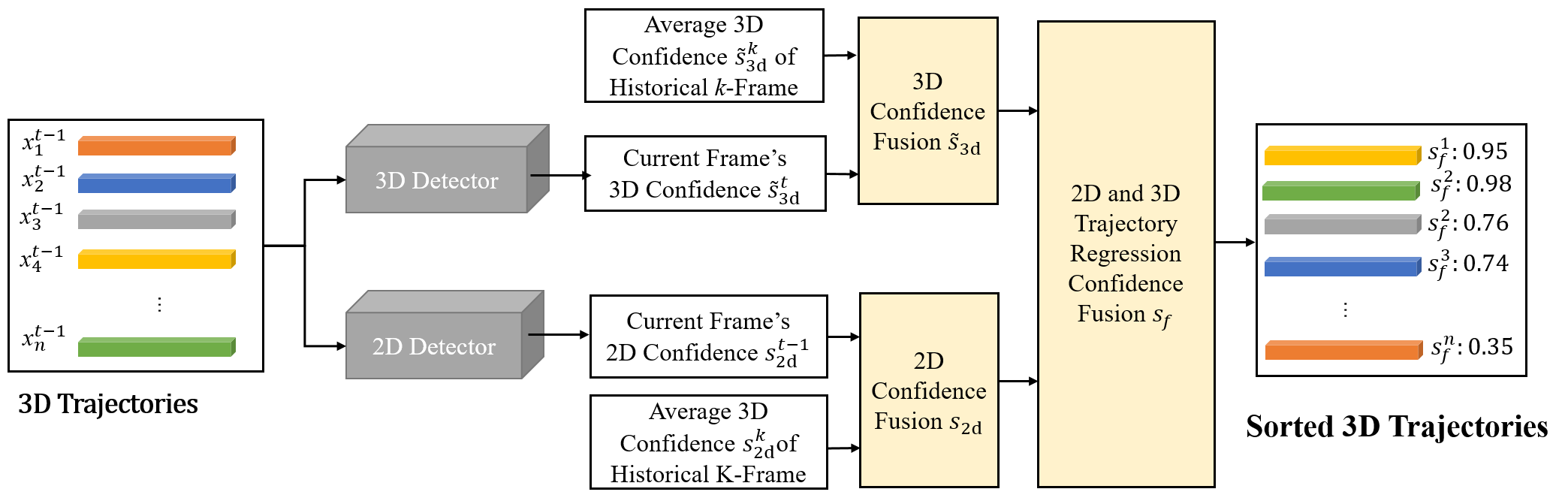}
	\caption{Schematic of the regression confidence fusion module.}
	\label{fig:fig4}
\end{figure}

As mentioned previously, “strong objects” and “weak objects” should be matched with different priority. Following this principle, all trajectories are sorted according to their regression confidence scores (i.e.  equation (\ref{eq:eq5}))), as shown in Figure \ref{fig:fig4}, and these sorted trajectories are then used for NMS to complete the tracking process.

\subsection{Post-processing of trajectories}

\quad This module mainly handles missed detections of the 3D detector (e.g. when dealing with remote objects), which leads to difficulties in matching historical trajectories and current detections. When using the 2D detector, remote objects can be more reliably detected. With the “post-processing” module on-board, if the regression confidence of historical 3D trajectories, projected to the 2D domain, exceeds the preset threshold, these trajectories are considered still existent and their states are set to “confirmed”.

	\section{Experiments}
\label{sec:experiments}
\subsection {Experimental Setup}

\quad The proposed MOT framework was implemented using the Python programming language on a desktop equipped with an AMD 7950X 4.5 GHz CPU, a 64 GB RAM, and two RTX 4090 graphics cards.

\textbf{Datasets}: Relevant data from the KITTI dataset \cite{13} and the Waymo dataset \cite{29} were used for experimental validation in this study. 

\textbf{Baselines}: To prove effectiveness of the proposed method, we compared our proposed method with the SOTA methods in the current literature. These baselines are given in Table \ref{tab:tab1}.

\textbf{Evaluation Metrics}: In this paper, the proposed method was evaluated using two important metrics CLEAR \cite{02} and HOTA \cite{20}. CLEAR is commonly used to evaluate the performance of multi-object tracking, and involves evaluation indicators such as multi-object tracking accuracy (MOTA), multi-object tracking precision (MOTP), ID switch (IDS). The CLEAR indicators are subdivided in the Waymo dataset into two levels, i.e. LEVEL\_1 (L1) and LEVEL\_2 (L2), based on the difficulty in object detection. L1 relates to easy-to-detect objects and L2 relates to difficult-to-detect objects. HOTA is currently used as a main evaluation method of tracking performance by the KITTI dataset, and it involves several indicators such as detection accuracy (DetA) and association accuracy (AssA). 

\titlespacing{\subsection}{0pt}{\parskip}{-\parskip}
\subsection {Tracking Performance Evaluation -- Quantitative Results}

\quad Our proposed method was first validated using the KITTI dataset, and the evaluation results are shown in Table \ref{tab:tab1}. These results were available at \url{https://www.cvlibs.net/datasets/kitti/eval\_tracking.php}.

\begin{table*}[t]
	\caption{Performance comparison with SOTA 3D MOT methods using the KITTI test sets. The best results are printed in bold. TBD refers to the tracking-by-detection paradigm, JDE refers to the joint detection and embedding paradigm, and JDT refers to the joint detection and tracking paradigm. }
	\begin{adjustbox}{max width=\textwidth}
		\begin{tabular}{cccccccccc}
			\toprule
			&  &  &  & \textbf{HOTA} & \textbf{AssA} & \textbf{MOTA} & \textbf{MOTP} &  &  \\
			\multirow{-2}{*}{\textbf{Method}} & \multirow{-2}{*}{\textbf{Published}} & \multirow{-2}{*}{\textbf{Type}} & \multirow{-2}{*}{\textbf{Input}} & \textbf{(\%)↑} & \textbf{(\%)↑} & \textbf{(\%)↑} & \textbf{(\%)↑} & \multirow{-2}{*}{\textbf{IDSW↓}} & \multirow{-2}{*}{\textbf{FN↓}} \\
			\rowcolor[HTML]{E7E6E6} 
			\midrule
			Quasi-Dense \cite{23} & 2021 CVPR & TBD & 2D & 68.45 & 65.49 & 84.93 & 84.85 & 313 & 3793 \\
			\rowcolor[HTML]{E7E6E6} 
			StrongSORT \cite{10} & 2023 TMM & TBD & 2D & 77.75 & 82.2 & \textbf{90.35} & 85.42 & 440 & --- \\
			\rowcolor[HTML]{E7E6E6} 
			LGM \cite{31} & 2021 ICCV & TBD & 2D & 73.14 & 72.31 & 87.6 & 84.12 & 448 & 2249 \\
			\rowcolor[HTML]{E7E6E6} 
			OC-SORT \cite{05} & 2023 CVPR & TBD & 2D & 76.54 & 76.39 & 90.28 & 85.53 & 250 & 2685 \\
			\rowcolor[HTML]{D9E2F3} 
			PC3TMOT \cite{39} & 2021 TITS & TBD & 3D & 77.8 & 81.59 & 88.81 & 84.26 & 225 & 2810 \\
			\rowcolor[HTML]{D9E2F3} 
			EAFFMOT\cite{17} & 2024 Signal   Processing & TBD & 3D & 72.28 & 73.08 & 84.77 & 85.08 & 107 & 3946 \\
			\rowcolor[HTML]{D9E2F3} 
			PolarMOT \cite{18} & 2022 ECCV & TBD & 3D & 75.16 & 76.95 & 85.08 & 85.63 & 462 & 2668 \\
			\rowcolor[HTML]{D9E2F3} 
			TripletTrack \cite{21} & 2022 CVPR & TBD & 3D & 73.58 & 74.66 & 84.32 & 86.06 & 322 & 4642 \\
			\rowcolor[HTML]{FBE4D5} 
			EagerMOT \cite{19} & 2021 ICRA & TBD & 2D+3D & 74.39 & 74.16 & 87.82 & 85.69 & 239 & 3497 \\
			\rowcolor[HTML]{FBE4D5} 
			DeepFusionMOT \cite{34} & 2022 RA-L & TBD & 2D+3D & 75.46 & 80.06 & 84.64 & 85.02 & 84 & 4601 \\
			\rowcolor[HTML]{FBE4D5} 
			StrongFusion-MOT \cite{33} & 2022 IEEE Sensors   Journal & TBD & 2D+3D & 75.65 & 79.84 & 85.53 & 85.07 & 58 & 4658 \\
			\rowcolor[HTML]{F7CAAC} 
			JMODT \cite{16} & 2021 IROS & JDE & 2D+3D & 70.73 & 68.76 & 85.35 & 85.37 & 350 & 3438 \\
			\rowcolor[HTML]{F7CAAC} 
			BcMODT \cite{41} & 2023 Remote   Sensing & JDE & 2D+3D & 71 & 69.14 & 85.48 & 85.31 & 381 & 3353 \\
			\rowcolor[HTML]{FFFFFF} 
			\midrule
			YONTD-MOT (Ours) & ---- & JDT & 2D+3D & \textbf{79.26} & \textbf{84.58} & 86.55 & \textbf{86.1} & \textbf{43} & \textbf{1784}    \\
			\bottomrule 
		\end{tabular}
	\end{adjustbox}
	\label{tab:tab1}
\end{table*}

Table \ref{tab:tab1} shows the tracking performance of recent SOTA MOT methods. It is seen in this table that the proposed method presents significant advantages in terms of several important metrics, including HOTA, DetA, AssA, MOTP, and IDSW. In terms of HOTA, 79.26\% was achieved by the proposed method, having improved by 3.8\% compared to DeepFusionMOT. For MOTP, an indicator focusing on the positional accuracy of trajectory boxes, the highest score of 86.10\% was achieved by our proposed method.  For AssA, an indicator reflecting the accuracy of association, our proposed method achieved a score of 84.58\%. Note that this score is a significant improvement compared to other latest MOT methods based on multi-modal fusion, such as DeepFusionMOT (80.06\%), JRMOT (66.89\%), and EagerMOT (74.16\%). In addition, the proposed method in this paper provided the least number of ID switches (IDSW), i.e. only 43 times, which demonstrates its robustness.

Apart from the commonly used KITTI dataset, another large dataset – the Waymo dataset – was also used for further evaluation. This dataset comprises a total of 3 categories. Comparative evaluation results are shown in Table \ref{tab:tab2}, and these performance data were sourced from submissions to the Waymo dataset website. For more details, please visit 
\url{https://Waymo.com/open/challenges/2020/3d-tracking/}.

We see in Table \ref{tab:tab2} that the proposed method also performed well on the Waymo dataset, having delivered the best tracking performance among all methods in comparison. Specifically, using our proposed method, significant improvement was achieved in multiple metrics compared with the Waymo baseline method – PK-baseline \cite{29}. For example, we observed improvements of 30.15\% in terms of MOTA (L1), 29.78\% in terms of MOTA (L2), 6.50\% in terms of MOTP (L1), and 6.54\% in terms of MOTP (L2). Compared with other SOTA methods, such as Probabilistic 3DMOT \cite{06}, our proposed method produced significant improvements in metrics such as MOTA (L1/L2) and MOTP (L1/L2).

\begin{table*}[h]
	\caption{Performance comparison with SOTA 3D MOT methods using the  waymo test sets. The best results are printed in bold.}
	\begin{adjustbox}{max width=\textwidth}
		\begin{tabular}{cccccccc}
			\toprule
			\multirow{2}{*}{\textbf{Method}} & \multirow{2}{*}{\textbf{Published}} & \textbf{MOTA/L1} & \textbf{MOTP/L1} & \textbf{FP/L1} & \textbf{MOTA/L2} & \textbf{MOTP/L2} & \textbf{FP/L2} \\
			&  & \textbf{(\%)↑} & \textbf{(\%)↑} & \textbf{(\%)↓} & \textbf{(\%)↑} & \textbf{(\%)↑} & \textbf{(\%)↓} \\
			\midrule
			PK-baseline \cite{29} & CVPR & 27.13 & 17.53 & 9.78 & 25.92 & 17.53 & 9.32 \\
			PV-RCNN-KF \cite{26} & CVPR & 57.14 & 24.95 & 8.72 & 55.53 & 24.97 & 8.66 \\
			2SDA-MOT \cite{08} & Sensors & 37.91 & \textbf{26.26} & 9.18 & 36.53 & \textbf{26.26} & 8.85 \\
			Probabilistic 3DMOT \cite{06} & ICRA & 49.16 & 24.8 & 9.13 & 47.65 & 24.82 & 8.99 \\
			YONTD-MOT (Ours) & ---- & \textbf{57.28} & 24.03 & \textbf{8.24} & \textbf{55.7} & 24.07 & \textbf{8.44}  \\
			\bottomrule 
		\end{tabular}
	\end{adjustbox}
	\label{tab:tab2}
\end{table*}

\subsection  {Tracking Performance Evaluation -- Qualitative Results}

\begin{figure*}[ht]
	\centering
	\includegraphics[width=\linewidth]{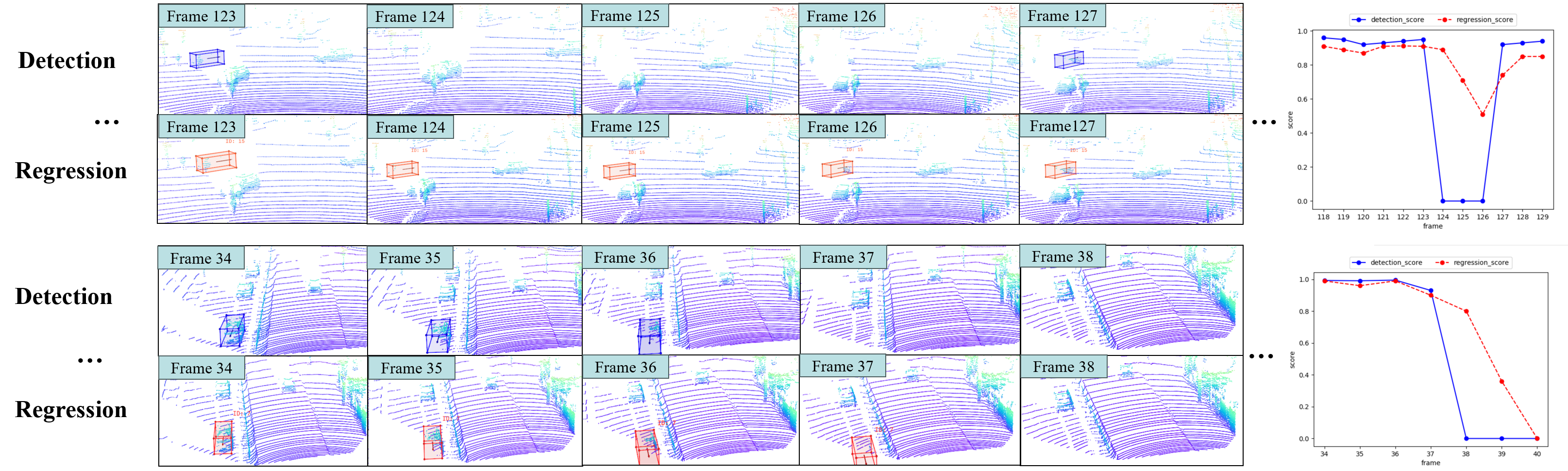}
	\caption{A visual comparison between the proposed trajectory regression confidence and the conventional detection confidence. By integrating historical trajectory features, our proposed trajectory regression confidence better reflects the actual patterns of objects’ motion changes over time.}
	\label{fig:fig5}
\end{figure*}

\quad Figure. \ref{fig:fig5} reveals that by using the trajectory regression confidence integrated with historical features, the actual motions of objects in the real world can be more genuinely reflected. The upper example in Figure. \ref{fig:fig5} shows that in the course of an object’s gradual occlusion and reappearance, its corresponding detection confidence undergoes drastic changes. Specifically, the detection confidence curve (in blue color) suddenly drops from a high confidence value to zero as occlusion occurs, and then sharply rises back up once the object reappears in the scene. In comparison, the trajectory regression confidence curve (in red color) gradually decreases (but not down to zero) as the situation of occlusion deteriorates, and then it gradually increases after occlusion disappears. Apparently, the variation pattern of this regression confidence curve is more consistent with the actual situation of the object’s occlusion and reappearance. The lower example in Figure. \ref{fig:fig5}  shows that when an object gradually disappears from the FOV, its regression confidence gradually reduces to zero, consistent with its actual pattern of motion. In contrast, the curve of the conventional detection confidence abruptly drops to zero immediately after the object disappears.

\begin{figure*}[ht]
	\centering
	\includegraphics[width=0.85\linewidth]{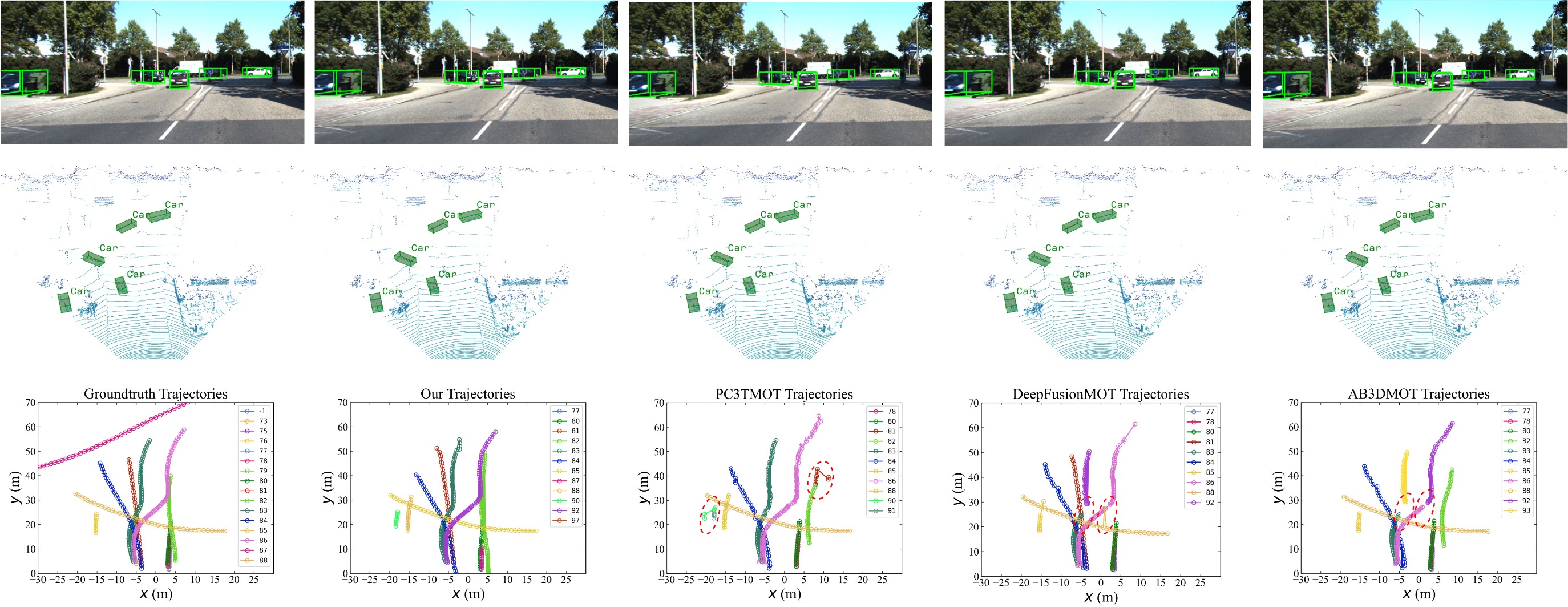}
	\caption{A comparison of bird’s eye view trajectories between our method and other SOTA methods including PC3TMOT, DeepFusionMOT, and AB3DMOT using a KITTI training sequence. Trajectories with the same ID are shown in the same color. The red circle in the figure indicates that there exists an ID switch in the trajectory.}
	\label{fig:fig6}
\end{figure*}

\begin{figure*}[ht]
	\centering
	\includegraphics[width=0.85\linewidth]{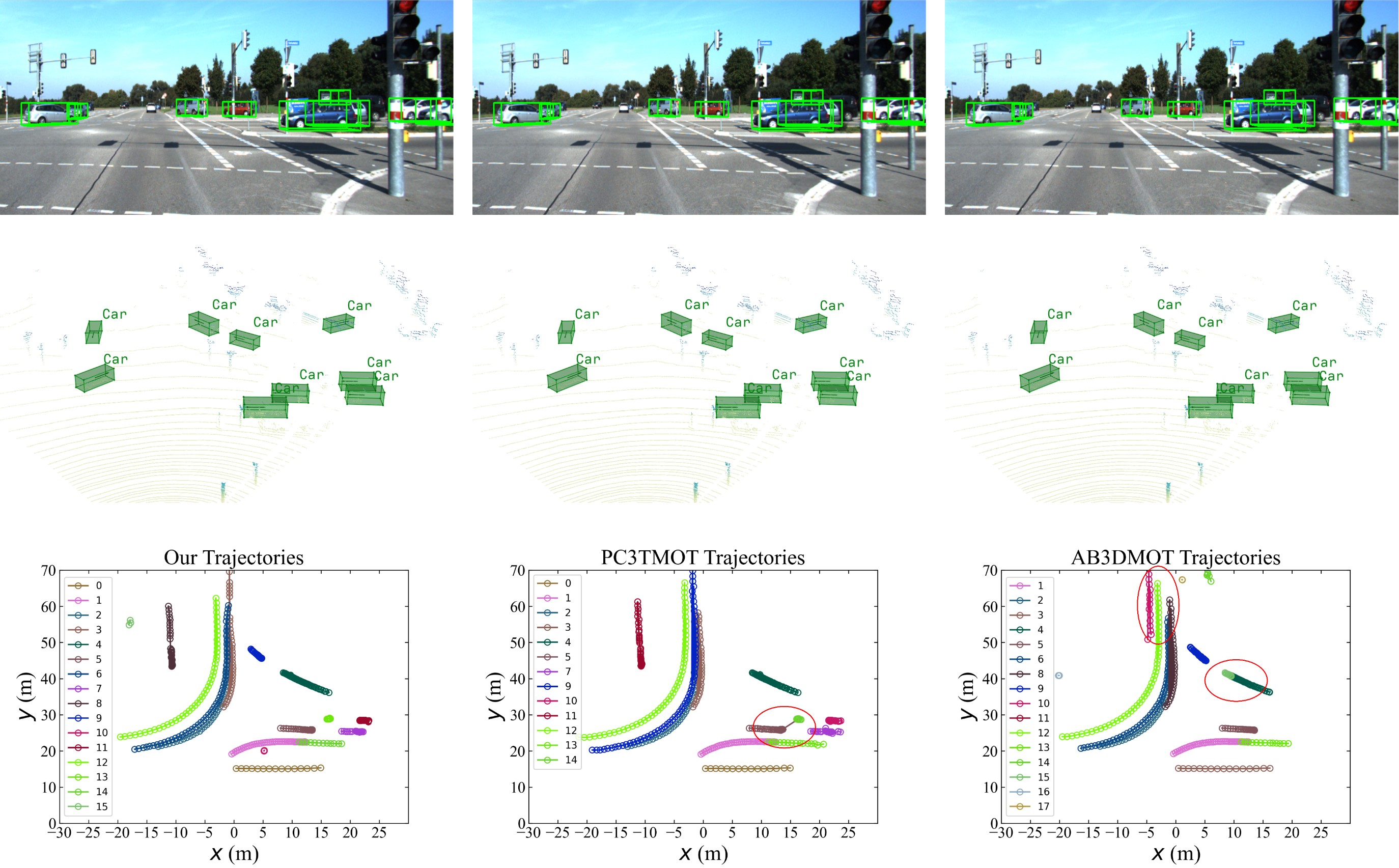}
	\caption{Another example of visual comparison between our method and other methods. The red circle in the figure indicates that there is an ID switch in the trajectory.}
	\label{fig:fig7}
\end{figure*}

Figure. \ref{fig:fig6} demonstrate a visual example of comparison with the latest trackers. Figure. \ref{fig:fig6} shows the KITTI training sequence 0001, where the first and second rows represent the detection results obtained in a certain frame. In the third row, the first column represents the ground truth trajectory, the second column denotes the result of the proposed method, the third column indicates the result of PC3TMOT, the fourth column stands for the result of DeepFusionMOT, and the fifth column is the result of AB3DMOT. We observe that using the PC3TMOT method, ID switches have occurred to trajectories with ID 90 and ID 91, and the same situation happened to trajectories with ID 81 and ID 82. Similarly, ID switches are also observed in the results of DeepFusionMOT and AB3DMOT. In contrast, by incorporating historical features when regressing historical trajectories, our proposed method demonstrates more robust performance in complex scenarios involving object occlusions and disappearances at intersections, without incurring any ID switches. Figure. \ref{fig:fig7} shows 0010 sequence in the KITTI test set, corresponding to a crossroad scene with object occlusion. In this circumstance, the proposed method also shows more robust performance compared to AB3DMOT and PC3TMOT. the method proposed in this paper displays more robust tracking performance.

\subsection  {Ablation Study}

\quad To explore the impacts of different modules on the overall tracking performance, ablation experiments were conducted using the validation sets in the KITTI datasets. Performance valuation was performed using two important metrics – CLEAR and HOTA.

\textbf {Impact of Object Detector Combinations}. To investigate the impact of different detector combinations on the overall tracking performance, three 3D detectors (i.e. CasA \cite{38}, PointRCNN \cite{27}, and Voxel R-CNN \cite{09}) and two 2D detectors (i.e. Faster R-CNN \cite{25} and Mask R-CNN \cite{15}) were selected in this ablation study. Experimental results are shown in Table \ref{tab:tab3}. We see that when Voxel R-CNN and Faster R-CNN were used, the highest HOTA, the best MOTA, and the least FN could be achieved. In contrast, combined use of PointRCNN and Mask R-CNN delivered the worst results.

\begin{table}[htpb]
	\centering
	\caption{Ablation study - impact of different object detector combinations on tracking performance.}
	\resizebox{\linewidth}{!}{
		\begin{tabular}{ccccc}
			\toprule
			\multirow{2}{*}{\textbf{3D Detector}} & \multirow{2}{*}{\textbf{2D Detector}} & \textbf{HOTA} & \textbf{MOTA} & \multirow{2}{*}{\textbf{FN↓}} \\
			&  & \textbf{(\%)↑} & \textbf{(\%)↑} &  \\
			\midrule
			\multirow{2}{*}{CasA \cite{38}} & Faster R-CNN \cite{25} & 76.37 & 80.41 & 1424 \\
			& Mask R-CNN \cite{15} & 75.3 & 77.81 & 1667 \\
			\multirow{2}{*}{PointRCNN \cite{27}} & Faster R-CNN \cite{25} & 53.11 & 43.96 & 4721 \\
			& Mask R-CNN \cite{15} & 55.02 & 46.59 & 4495 \\
			\multirow{2}{*}{Voxel R-CNN \cite{09}} & Faster R-CNN \cite{25} & \textbf{77.52} & \textbf{82.11} & \textbf{610} \\
			& Mask R-CNN \cite{15} & 76.5 & 80.36 & 786  \\
			\bottomrule 
	\end{tabular}}
	\label{tab:tab3}
\end{table}

\textbf {Impact of Historical Features}. In this ablation study, Voxel R-CNN was employed as the 3D detector and Faster RCNN was used as the 2D detector. We see in Table \ref{tab:tab4} that the tracking performance was enhanced by integrating the 3D features of historical trajectories and feeding them into the 3D detector for regression. However, for the Voxel R-CNN detector, incorporating longer temporal features does not necessarily lead to better results. Instead, blending only one frame of historical information provides the best outcome. The reason why fusing the features of one frame works best is that we did not retrain the detector. The original detector was trained using single-frame features. We believe that for moving objects, the features in consecutive frames are quite similar, while incorporating features from a longer time span may confuse a detector that has not been retrained.

\vspace{5pt}
\begin{table}[htpb]
	\centering
	\caption{The impact of the number of fused historical trajectory feature frames on tracking performance.}
	\resizebox{\linewidth}{!}{
		\begin{tabular}{cccccc}
			\toprule
			\multirow{2}{*}{\textbf{\begin{tabular}[c]{@{}c@{}}Historical \\ Features\end{tabular}}} & \multirow{2}{*}{\textbf{\begin{tabular}[c]{@{}c@{}}Historical \\ Frames\end{tabular}}} & \textbf{HOTA} & \textbf{MOTA} & \multirow{2}{*}{\textbf{FN↓}} & \multirow{2}{*}{\textbf{FP↓}} \\
			&  & \textbf{(\%)↑} & \textbf{(\%)↑} &  &  \\
			\midrule
			\textbf{×} & \textbf{×} & 77.52 & 82.11 & 610 & 1135 \\
			\checkmark  & 1 & \textbf{78.34} & \textbf{83.26} & \textbf{575} & \textbf{1056} \\
			\checkmark  & 5 & 77.68 & 82.86 & 602 & 1069 \\
			\checkmark  & 30 & 77.55 & 82.32 & 611 & 1113 \\
			\checkmark  & ALL & 77.43 & 82.05 & 609 & 1138   \\
			\bottomrule
	\end{tabular}}
	\label{tab:tab4}
\end{table}  

\textbf {Impact of 2D Detector}.The MOT framework proposed in this paper can achieve high-precision tracking with a 2D detector. Removal of the 2D detector downgrades the framework to a point cloud-based end-to-end multi-object tracker, and thereby resulting in deterioration of tracking performance. As shown in Table \ref{tab:tab5}, removal of the 2D detector leads to significant exacerbation in metrics such as HOTA, and MOTA, mainly due to the significant increase in FP. The reason for the increase of FP, after removing the 2D detector, is that the 3D detector experiences significant performance deterioration for detecting remote objects, as point clouds reflected by remote objects are very sparse. In comparison, the 2D detector works based on image pixels, which are significantly less sensitive to distance increase. It has been shown in DeepFusionMOT and EagerMOT that 2D detectors outperform 3D detectors in terms of detecting remote objects. In summary, the 2D detector provides more accurate object detection information for the proposed method, thereby improving accuracy and robustness of the tracker.

\vspace{5pt}
\begin{table}[htbp]
	\centering
	\caption{Ablation study - impact of 2D detector on tracking performance.}
	\resizebox{\linewidth}{!}{
		\begin{tabular}{ccccccc}
			\toprule
			\multicolumn{2}{c}{\textbf{Detector}} & \multirow{2}{*}{\textbf{\begin{tabular}[c]{@{}c@{}}Historical \\ Features\end{tabular}}} & \multirow{2}{*}{\textbf{\begin{tabular}[c]{@{}c@{}}HOTA\\ (\%)↑\end{tabular}}} & \multirow{2}{*}{\textbf{\begin{tabular}[c]{@{}c@{}}MOTA\\ (\%)↑\end{tabular}}} & \multirow{2}{*}{\textbf{FN↓}} & \multirow{2}{*}{\textbf{FP↓}} \\
			\textbf{3D} & \textbf{2D} &  &  &  &  &  \\
			\midrule
			\checkmark & \textbf{×} & \checkmark & 71.16 & 69.63 & 696 & 2279 \\
			\checkmark & \checkmark & \checkmark & \textbf{78.34} & \textbf{83.26} & \textbf{575} & \textbf{1056}   \\
			\bottomrule
	\end{tabular}}
	\label{tab:tab5}
\end{table}

\textbf {Impact of NMS order}.
To investigate the impact of NMS order of trajectories and detections (i.e. the impact of the proposed confidence fusion module) on tracking performance, three sets of experiments were conducted using the above-mentioned KITTI validation set: unordered NMS, NMS in descending order of confidence, and NMS in ascending order of confidence. Experimental results are shown in Tables \ref{tab:tab6}. We see in Table \ref{tab:tab6} that when the trajectory regression confidence was sorted in descending order, best results of important metrics including HOTA, AssA, and MOTA were achieved. The experimental results above suggest that the confidence fusion module proposed in this paper effectively improves tracking performance.

\vspace{5pt}
\begin{table}[htbp]
	\centering
	\caption{Ablation study - impact of different NMS orders.}
	\resizebox{\linewidth}{!}{
		\begin{tabular}{ccccccc}
			\toprule
			\multirow{2}{*}{\textbf{\begin{tabular}[c]{@{}c@{}}Confidence \\ order\end{tabular}}} & \multirow{2}{*}{\textbf{\begin{tabular}[c]{@{}c@{}}2D \\ Detector\end{tabular}}} & \multirow{2}{*}{\textbf{\begin{tabular}[c]{@{}c@{}}Historical \\ Features\end{tabular}}} & \multirow{2}{*}{\textbf{\begin{tabular}[c]{@{}c@{}}HOTA \\ (\%)↑\end{tabular}}} & \multirow{2}{*}{\textbf{\begin{tabular}[c]{@{}c@{}}AssA \\ (\%)↑\end{tabular}}} & \multirow{2}{*}{\textbf{\begin{tabular}[c]{@{}c@{}}MOTA \\ (\%)↑\end{tabular}}} & \multirow{2}{*}{\textbf{FP↓}} \\
			&  &  &  &  &  &  \\
			\midrule
			Unordered & \checkmark & \checkmark & 77.6 & 81.88 & 82.75 & 1101 \\
			Ascending & \checkmark & \checkmark & 74.38 & 79.34 & 75.41 & 1706 \\
			Descending & \checkmark & \checkmark & \textbf{78.34} & \textbf{83.01} & \textbf{83.26} & \textbf{1056}  \\
			\bottomrule
	\end{tabular}}
	\label{tab:tab6}
\end{table}  

	\section{Conclusion}
\label{sec:conclusion}

\quad In this paper, a new MOT framework was proposed based on multimodal fusion. This framework achieves good tracking performance without the need for complex data association and outperforms many SOTA multimodal tracking methods that employ various features for association. In our proposed approach, for the first time, the trajectory regression confidence is used to characterize the trajectory state (strong object or weak object), and trajectories are sorted for ordered tracking. Evaluation of the proposed method was conducted using the KITTI and Waymo datasets. Results indicate that SOTA performance was achieved with the KITTI dataset compared to current multimodal fusion-based methods on the leaderboard, and good tracking performance was also achieved with the Waymo dataset.

	
	
\end{document}


\title{\paperTitle}
\author{\authorBlock}
\maketitlesupplementary

\section{Appendix Section}
Supplementary material goes here.

{\small
\bibliographystyle{ieee_fullname}
\bibliography{11_references}
}